\documentclass[10pt,twocolumn,letterpaper]{article}

\usepackage{iccv}
\usepackage{times}
\usepackage{epsfig}
\usepackage{graphicx}
\usepackage{amsmath}
\usepackage{amssymb}
\usepackage{subfigure}
\usepackage{color}
\usepackage{multirow}
\usepackage{textcomp, gensymb}
\usepackage{bbding}
\usepackage{booktabs}
\usepackage{bm, eucal}
\usepackage{pifont}
\usepackage{enumitem}

\usepackage{caption, multirow, overpic, textpos}
\usepackage{booktabs}
\usepackage[table]{xcolor}

\usepackage{xspace}
\usepackage{balance}
\usepackage{numprint}
\usepackage{sidecap}
\usepackage{float}
\usepackage{siunitx}
\usepackage{pifont}

\usepackage[pagebackref=true,breaklinks=true,letterpaper=true,colorlinks,bookmarks=false]{hyperref}

\iccvfinalcopy 

\def\iccvPaperID{6079} 

\ificcvfinal\fi

\begin{document}

\title{Learning Dense UV Completion for Human Mesh Recovery}

\author{Yanjun Wang$^{1}$ \quad \; Qingping Sun$^{2}$ \quad \; Wenjia Wang$^3$ \quad \; Jun Ling$^1$ \quad \; \\ Zhongang Cai$^2$ \quad  \; Rong Xie$^1$ \quad \; Li Song$^{1, \dagger}$\\[1.5mm]
\normalsize $^1$ Shanghai Jiao Tong University \quad
\normalsize $^2$ Sensetime Research \quad
\normalsize $^3$ Shanghai AI Laboratory \\
}

\ificcvfinal\thispagestyle{empty}\fi

\twocolumn[{%
	\renewcommand\twocolumn[1][]{#1}%
	\maketitle
	\begin{center}
		\newcommand{\teaserwidth}{0.8\linewidth}
		\centerline{
			\includegraphics[width=0.8\linewidth,clip]{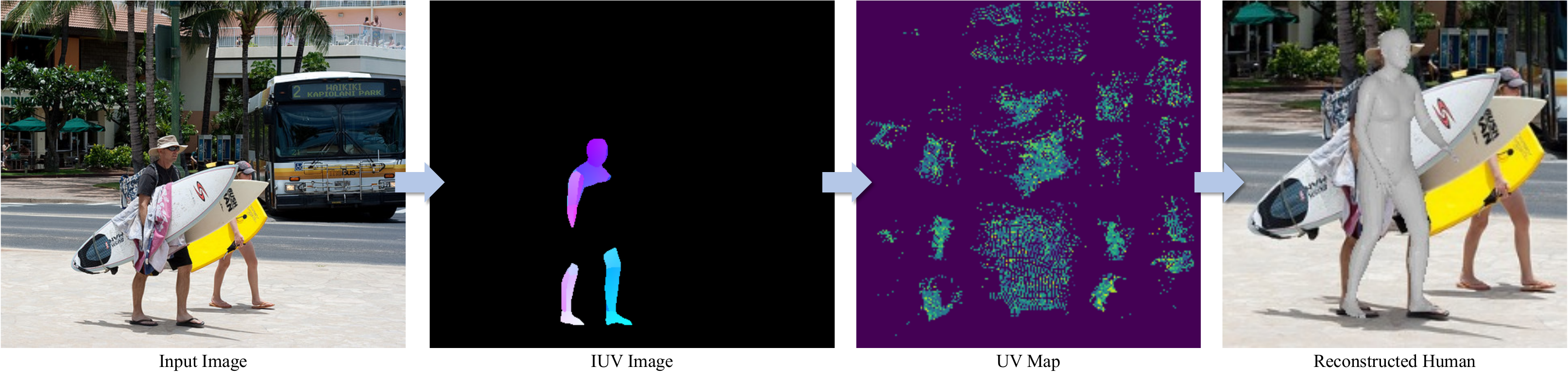}
		}
		\vspace{-1ex}
		\captionof{figure}{\label{fig:teaser}
Our method consists of two key steps: 1) estimates the dense UV map of the obscured body and mask out the occlusion regions; 2) performs UV-based feature wrapping and completion for future mesh reconstruction. 
  }
	\end{center}%
}]

{
  \renewcommand{\thefootnote}%
    {\fnsymbol{footnote}}
  \footnotetext[1]{LS$^\dagger$ is the corresponding author.}
}


\begin{abstract}
Human mesh reconstruction from a single image is challenging in the presence of occlusion, which can be caused by self, objects, or other humans. Existing methods either fail to separate human features accurately or lack proper supervision for feature completion. In this paper, we propose Dense Inpainting Human Mesh Recovery (DIMR), a two-stage method that leverages dense correspondence maps to handle occlusion. Our method utilizes a dense correspondence map to separate visible human features and completes human features on a structured UV map dense human with an attention-based feature completion module. We also design a feature inpainting training procedure that guides the network to learn from unoccluded features. We evaluate our method on several datasets and demonstrate its superior performance under heavily occluded scenarios compared to other methods.
Extensive experiments show that our method obviously outperforms prior SOTA methods on heavily occluded images and achieves comparable results on the standard benchmarks (3DPW).
\end{abstract}

\section{Introduction}
\label{sec:intro}


Reconstructing human mesh from a single image broadens a broad view of applications, such as human motion analysis, digital human animation, augmented reality, and human-world interactions. However, occlusion is a long-standing obstacle that hinders the model’s performance for in-the-wild images. Occlusion can be divided into three categories: self-occlusion, the occlusion of certain parts caused by the body itself; object occlusion, the occlusion caused by other objects in the environment; inter-human occlusion, the occlusion caused by closely overlapped humans. Among them, self and object occlusion can all be seen as non-human occlusion, as they only lack information rather than have confusing information introduced by other humans.

The widespread parametric model SMPL~\cite{smpl,smplx} has facilitated rapid progress in this field in recent years, with various approaches proposed to improve the model generalization performance. Most of them aim at completing the missing features caused by occlusion. HMR~\cite{hmr} proposes to use an adversarial prior to correct the implausible poses. This can be seen as a completion from the pose prior. But the distribution of the prior largely depends on its dataset. Thus, the variety of the poses is limited by the conclusiveness of the prior datasets. Other methods seek cues from the image feature. PARE~\cite{pare} proposes a part-attention mechanism, which can utilize local features from visible body parts to complete the occluded body's feature. But for these image-based methods, the completion method lacks intermediate supervision, and their performance is often limited.

Human occlusion presents a significant challenge compared to non-human occlusion because it provides ambiguous information that can lead to difficulties in distinguishing between different body parts, such as limbs and torso. In response to this challenge, several methods have been proposed to model multiple humans. ROMP~\cite{romp} and BEV~\cite{bev} are two representative multi-person methods that regress multiple persons directly. They represent humans with 2D or 3D heat maps and solve the position ambiguity of humans in various ways. However, due to limitations in resolution, these methods may not accurately regress every person, and they may not have solutions for occluded body parts. Two-stage methods designed to predict one person at a time tend to focus on separating and identifying the true human to predict. OCHMR~\cite{ochmr} utilizes a ROMP-liked center heat map. 3DCrowdNet\cite{3dcrowdnet} uses a keypoint heatmap. Pose2UV~\cite{pose2uv} combines segmentation with a keypoint heatmap for separation. But these heatmap or segmentation cues are only used as an auxiliary input and guide the network implicitly. These cues cannot clearly separate the predicted and occluded humans, which might introduce confusing features when two body parts are close enough.

In this work, we resort to solving general occlusion while maintaining relatively high performance. More specifically, a model takes as input a high-resolution image and has the ability to complete missing features with ground truth supervision. For the separation issue, we draw inspiration from the dense corresponding map, which regresses each pixel into a surface point on the human model. Previous studies~\cite{decomr,visdb} prove that dense correspondence is beneficial for human reasoning. But their dense prediction is based on the image input, and weakly supervised, which means the prediction itself will be affected by occlusion. For the completion issue, OOH~\cite{ooh} utilizes a dense location map for direct supervision of the occluded vertices through inpainting. Compared to SMPL parameters, this approach provides much denser and more direct supervision. However, direct regression of vertices coordinates using this method requires more regulation to achieve a smooth surface.

Inspired by these methods utilizing dense correspondence, we propose Dense Inpaiting Human Mesh Recovery (DIMR), a top-down, high-resolution method designed to solve different occlusions. To leverage the representative power of dense correspondence maps to separate human features, we utilize a dedicated network for dense correspondence map regression. The regressed dense map is much more accurate and sharp compared to an integrated module. Moreover, the network simultaneously performs human detection and dense map prediction, which is suited to the first stage of our module. But utilizing the dense map for parameter regression and occlusion handling is non-trivial. Given the UV nature of the dense map, we designed a feature-wrapping process to accurately separate human features and project them to a structured UV map. Based on this UV map, we further proposed an attention-based mechanism to complete the occluded feature based on the other part's features. And based on the attention, we derived part-wise human features to regress SMPL parameters. To further enhance the occlusion handling ability, we designed a feature inpainting training procedure that guides the network for feature completion with the unoccluded features. To validate our design, we conduct extensive experiments on the 3DOH~\cite{ooh},  3DPW~\cite{3dpw}, 3DPW-OC~\cite{3dpw,pare} and 3DPW-PC~\cite{3dpw}  datasets. The results indicate that our model yields a higher accuracy under heavily occluded scenarios compared to other occlusion handling experiments and maintains a comparable accuracy on general human datasets.

In summary, our contributions are three folds:
\begin{itemize}
    \item We introduce a novel way to utilize a dense correspondence map that learns human features reconstruction under heavy inter-person occlusion.
    \item We propose a mechanism utilizing visible features to complete occluded parts features to reason object occlusion. And designed a training mechanism with direct visible supervision to guide the completion.
    \item Our method achieves state-of-art results on occlusion datasets of 3DPW-PC, 3DPW-OC, and has a comparable accuracy on general datasets on 3DPW. Extensive experiments show that our method is capable of dealing with human mesh recovery under heavy occlusion.
\end{itemize}

\section{Related Work}

\subsection{Human Mesh Recovery}

Recovering 3D human pose and shape only from a single image is a challenging task due to the lack of multiview information and complicated human articulations. Most recent works can generally be divided into optimization-based methods and learning-based methods. Optimization-based methods~\cite{MuVS,up3d,guan2009estimating,smplify,fang2021mirrored} work by fitting a statistical human body model to the 2D cues extracted from the input image, such as 2D keypoints~\cite{smplify} or silhouette~\cite{up3d}. On the other hand, learning-based methods have gained significant traction in recent years and employ deep neural networks to estimate the 3D pose and shape of the human body, which can be further classified into two categories: model-based and model-free methods.

Model-based methods~\cite{hmr,spec,pare,pymaf,romp,bev,hybrik,tuch,shapy} leverage a deep neural network to regress the parameters of the human model. HMR~\cite{hmr} first utilizes CNN to extract the global features and then regresses the SMPL parameters. SPIN~\cite{spin} combines the learning-based method (HMR) with the optimization-based method (SMPLify), which leads to accurate image-model alignment. CLIFF~\cite{cliff} proposes a new camera system to solve the inaccuracy of global rotations under the cropped image. With a corrected camera system, it outperforms prior arts on a significant margin even with a simple HMR-liked structure.

Leverage the efficiency of 3D mesh reconstruction tasks~\cite{c-axis, global, rfeps,topdown, close}, model-free methods~\cite{graphcmr,i2l,pose2mesh,decomr,metro,meshgraphormer, fastmetro, tore} leverage a deep neural network to directly estimate the 3D coordinate of the human mesh. GraphCMR~\cite{graphcmr} is a milestone work, and it uses CNN to extract features from the input image at first and then uses GraphCNN~\cite{kipf2016semi} to estimate the 3D coordinates of the human mesh directly. Although GraphCNN considers local interactions among neighbor vertices, like HMR, GraphCMR only uses global features extracted from the image to recover the human mesh. 
Although the accuracy of unoccluded, single-person benchmarks have been increasing, these methods cannot generalize well to occlusion scenarios.

\subsection{Occlusion-aware Human Mesh Recovery}
Recent efforts that aim at solving occlusions mainly focus on two main groups: objects or human occlusions~\cite{bev,romp,ooh,pare,3dcrowdnet,jiang2020mpshape}

For object occlusion, PARE~\cite{pare} devises a part attention regressor to predict body-part-guided attention mask, which is helpful for the neural network to exploit information about the visibility of individual body parts to estimate occluded parts, but it overlooked the importance of global feature and the connection between different joints. 
OOH~\cite{ooh} proposes a method to complete the model from the surface prior. It maps the SMPL model to a UV image and masks out the pixel value of the occluded surface. It models occlusion completion as an image inpainting task, using the network to complete the occluded UV image value.

For inter-human occlusion, ROMP~\cite{romp} proposes a collision-aware center heatmap to model multiple humans on the image. It builds a repulsive field of body centers that push away close body centers, making overlapping human body centers easier to distinguish. Jiang et al.~\cite{singleshotmultiperson2018} utilize an interpenetration loss to regular multiperson location based on instance segmentation. 3DCrowdNet~\cite{ochmr} proposes to distinguish the target's image feature with the occluded human with the guide of a 2D joint map. Compared to the sparse representation of a predicted human, our model utilizes a dense correspondence map to achieve accurate and dense image feature separation.

\subsection{Dense Correspondence Learning}

Dense human body representations~\cite{alldieck2019tex2shape,decomr,visdb} have been widely used in the analysis of humans.  DecoMR~\cite{decomr} propose to estimate a dense correspondence map and wrapped the feature to $UV$ space for coordinate regression. VisDB~\cite{visdb} predicts 3D heatmaps for human joints and vertices and the visibility for each vertex. These methods prove that dense correspondence is beneficial for human reasoning. But their dense prediction is based on the image input, and weakly supervised, which means the prediction itself will be affected by occlusion. DensePose~\cite{Guler2018DensePose} proposed a CNN-based system that regresses coordinates of the human body surface on the human image, which can be later utilized for feature transfer, human mesh regressing. This dedicated design has much higher accuracy and is capable of separating the features of occluded persons. Our model leverages its ability to predict accurate dense correspondence images.

\section{Method}
\label{chapter:method}

DIMR utilizes a human dense correspondence map, also called the IUV image, to aggregate visible features of the occluded human as well as isolate the prediction subject from the overlapped human occlusion. This section begins with a brief overview of the SMPL~\cite{smpl} body model and subsequently discusses the design of our model. Based on the observation of current occlusion handling methods, we integrate the separation of human instances and the completion of human features with a dense IUV image.  To better separate human instances, we use the map to wrap the human feature to a specially designed UV map. Features are rearranged based on the body part they belong to. Given the structured UV feature. Also, we proposed an attention-based method to promote feature completion on the structured feature map. Finally, to better handle occlusion, we proposed a parallel training method to feed feature-level supervision to the network.

\subsection{Preliminary}

\noindent\textbf{SMPL } represents a 3D human mesh with $ \mathcal{M}(\boldsymbol{\theta}, \boldsymbol{\beta}) \in \mathbb{R}^{6890 \times 3}$ where $\boldsymbol{\beta} \in \mathbb{R}^{10}$  represents the first 10 coefficients for SMPL's PCA surface shape, and $\boldsymbol{\theta} \in \mathbb{R}^{24 \times 3}$ represents the joints rotations which include global rotations. Given the 3D mesh, the 3D joints can be calculated as $J_{3D} = \mathcal{J} \mathcal{M}\in \mathbb{R}^{J \times 3}, J=24$ with a pretrained joint regressor $ \mathcal{J} $. 

We adopt a weak perspective camera system $\boldsymbol{\pi}=(s, \textbf{t}), \textbf{t} \in \mathbb{R}^{2} $ to project the predicted body model to the image plane. Based on 3D joints and camera, we caculate the 2D joints following:  $J_{2 D}=s \Pi( \mathcal{J}\mathcal{M})+\textbf{t}  \in \mathbb{R}^{J \times 3} $, where $\Pi$ is orthographic projection.

\begin{figure*}[t]
  \centering
   \includegraphics[width=0.9\linewidth]{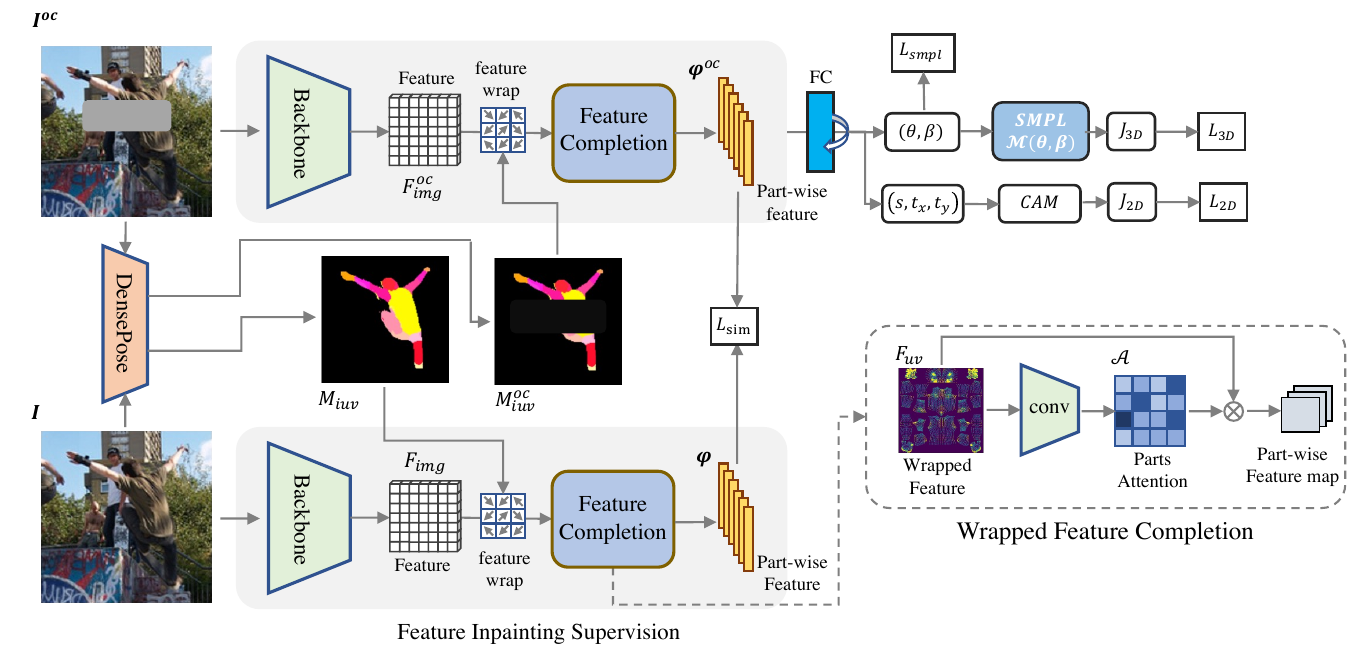}
  \caption{\textbf{Pipeline overview} Our model consists of a UV feature wrap module, a feature completion model, and an FC layer for SMPL parameter regression. Our network takes the image $I$ and predicted IUV image $M_{iuv}$ as input. The IUV image wraps the image feature to UV space. The attention-based feature completion module completes the occluded feature based on the visible ones. The inpainting training takes the original and occluded image as input, and uses the original feature as the supervision for the occluded feature.}
  \label{fig:pipeline}
\end{figure*}

\noindent\textbf{Input Preparation.} The pipeline of our model is shown in Figure~\ref{fig:pipeline}. Given an uncropped image, we first utilize DensePose~\cite{Guler2018DensePose} to extract the IUV map of every object. Based on each dense map and bounding box, our network takes the cropped image $I \in \mathbb{R}^{3 \times H \times W}$and cropped IUV image $M_{iuv} \in \mathbb{R}^{3 \times H \times W}$ as input. In $M_{iuv}$, the first channel $i\in \mathbb{R}^{H \times W}$ represents the segmentation of the human object, and the second and third channels are a two-dimensional coordinate $(u,v)\in \mathbb{R}^{2 \times H \times W}$, which represents the mapping location of this pixel on the UV map.

\subsection{UV Based Feature Wrapping}

We utilize an image encoder to encode the cropped image and obtain a feature map $\mathbf{F_{img}} \in \mathbb{R}^{D \times H^{\prime} \times W^{\prime}}$. The feature wrapping module wraps the feature vector on ${F_{img}}$ into a UV space feature map ${F_{UV} } \in \mathbb{R}^{D \times H^{\prime} \times W^{\prime}}$ according to the following equation:
 \begin{equation}
 \centering
 \begin{split}
 {F_{UV} }(u(x,y),v(x,y)) = F_{img}(x,y), i(x,y)>0\\
    x=0,1,...,(H^{\prime}-1), y=0,1,...,(W^{\prime}-1) 
\end{split}
\end{equation}

\begin{figure}[ht]
  \centering
    \centering
    \includegraphics[width=0.7\linewidth]{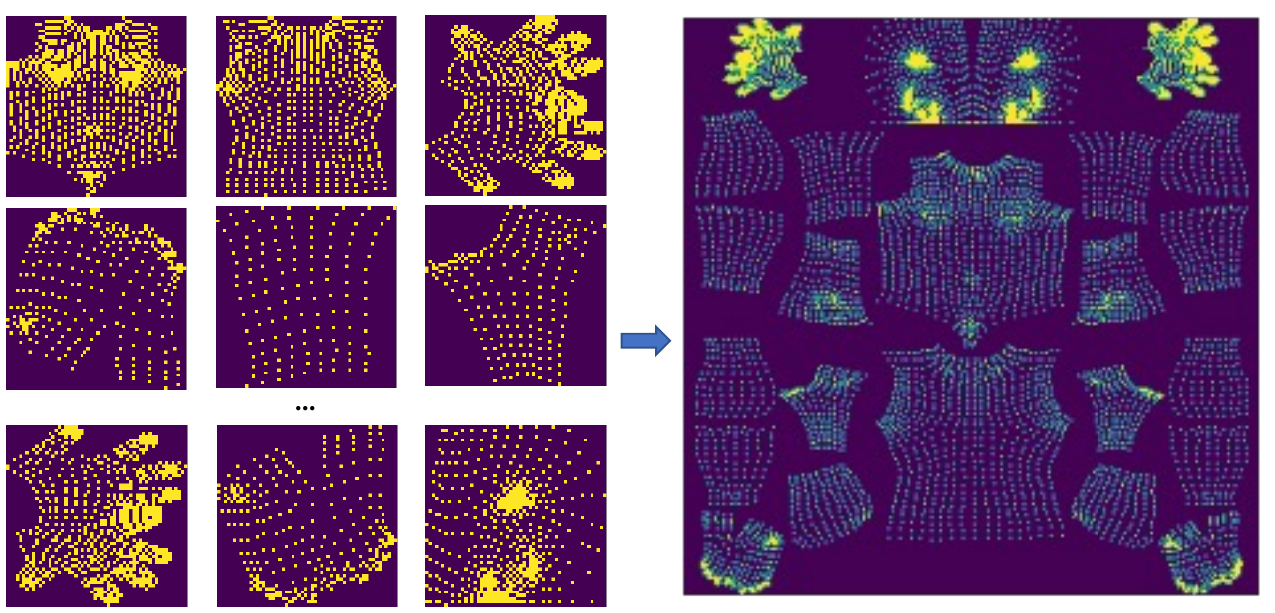}
    \caption{\textbf{Illustration of UV map rearrangement.} The original DensePose UV map(left) is part-based. But the density of mapped features varies dramatically. We rearrange the map together according to their actual size and relative position.}
    \label{fig:iuv_arrange}
    
\end{figure}

This procedure involves rearranging image features in a structured manner by clustering features belonging to each body part. As shown in Figure~\ref{fig:iuv_arrange}, the initial UV maps produced by DensePose are part-wise. But the part UV maps are imbalanced compared to a holistic UV map. The actual scale of different body parts varies, leading to a significant difference in the density of valid feature vectors across different maps. And the separation of different parts makes inter-part feature completion challenging. So we map the part-wise UV map into a single map, considering the relative size and position of different parts. We think this arrangement could preserve more neighboring body-part relationships. 

\subsection{Wrapped Feature Completion}


The wrapped feature completion module aims to utilize the visible features to reason the occluded features. The module first uses convolution blocks to downsample the wrapped feature map, and combine the neighboring features. This process enables the module to fuse the features within a body part and those around neighboring parts. To enable the module to consider features across the map, we propose an attention method to further refer to features from different body parts. It applies two $1 \times 1$ convolution layers to the downscaled feature map $F^{d}_{UV}$ to derive body part attention weight $\mathcal{A}$ and attention value $\mathcal{V}$. Then, it weighted sums the attention value through element-wise matrix multiplication formulated as:
\begin{equation}
\boldsymbol{\varphi}_i=\sum_{h,w}\sigma(\mathcal{A_i}) \circ \mathcal{V}
\end{equation}
where $i$ denotes $i^{th}$ body-part, $\circ$ denotes element-wise multiplication and $\sigma$ is softmax function that normalizes attention map. As a consequence of feature wrapping, the part feature of different instances is expected to be situated in the same location. Consequently, the attention map for a given part should emphasizes a fixed position as the primary reference for the part feature $\boldsymbol{\varphi}_i$. Additionally, since various instances may exhibit diverse occlusions, the attention map should also have several other focus areas to obtain information from other parts. It is noteworthy that we train this model without explicit supervision, as the location of the part features is expected to be similar.

\subsection{UV Based Feature Inpainting}

Using feature wrapping and feature completion module, we have obtained an accurate part-wise feature vector. Our goal is to ensure that these features are robust and invariant to occlusions. Previous methods~\cite{pare} have relied on data augmentation strategies, such as random cropping or synthetic occlusion, to achieve this goal. However, experiments conducted by Pang \emph{et. al.}~\cite{pang2022benchmarking} indicate that such augmentation methods may not be as helpful as previously believed. We recognize that it is challenging for the network to learn similar features when the image is under occlusion. With only SMPL parameter supervision, the model can hardly learn to output similar human features when the image is occluded.

To address this issue, we propose a feature inpainting training method that provides part-wise supervision for occlusion augmentation. As shown in Figure~\ref{fig:pipeline}, this training method takes two inputs, the original image $I$ and the occluded image $I^{occ}$ augmented by synthetic occlusion. We also add occlusion as a binary mask on the predicted IUV image, denoted as $M_{iuv}$ and $M_{iuv}^{occ}$. The network processes both pairs of inputs. The unoccluded part feature vector  $\boldsymbol{\varphi}$ serves as the supervision to regulate the occluded vector $\boldsymbol{\varphi}^{oc}$. The objective is to predict similar part features even under occlusion. We model this as a similarity loss formulated as:
\begin{equation}
    L_{sim}=\frac{1}{p} \sum_{i=0}^{p-1} \frac{\boldsymbol{\varphi}^{oc}_i \boldsymbol{\varphi}_i}{|\boldsymbol{\varphi}^{oc}_i||\boldsymbol{\varphi}_i|}
\end{equation}

This training strategy introduces the ground truth feature to the network, helping the model to inpaint the missing feature. 

The inpainted part feature $\boldsymbol{\varphi}_i$ is flattened and concatenated with global feature average pooled from $F_{img}$, initial SMPL and camera parameters $(\boldsymbol{\theta}, \boldsymbol{\beta}, \boldsymbol{\pi})_{init}$. The final fully-connected layer regresses the final parameters $(\boldsymbol{\theta}, \boldsymbol{\beta}, \boldsymbol{\pi})$ in an iterative regression manner. 

\subsection{Loss Functions}
We superviseF our training with SMPL parameters loss $L_{SMPL}$, 3D keypoints loss $L_{3D}$,  2D keypoints $L_{2D}$ and similarity loss $L_{sim}$ that can be formulated as:
\begin{equation}
 \begin{aligned}
L =\lambda_{S M P L} L_{S M P L}+\lambda_{3 D} L_{3 D}+\lambda_{2 D} L_{2 D}
+\lambda_{sim} L_{sim}
\end{aligned}
\end{equation}
\begin{equation}
\begin{aligned}
L_{S M P L} =& \|(\boldsymbol{\theta}, \boldsymbol{\beta})-(\boldsymbol{\hat\theta}, \boldsymbol{\hat\beta)}\|, \\
L_{3 D} =& \left\|J_{3 D}-\hat{J}_{3 D}\right\|, \\
L_{2 D} =& \left\|J_{2 D}-\hat{J}_{2 D}\right\|,
\end{aligned}
\end{equation}

\subsection{Implementation Details}
We utilize HRNet \cite{cheng2020higherhrnet} as the image encoder.  Compared to ResNet\cite{resnet}, HRNet's design preserves more image-aligned features and has a larger resolution feature map output, which makes the wrapped feature more precise and corresponds to the image pixel. We set 

We implement DIMR with PyTorch~\cite{NEURIPS2019_9015} and MMHuman3D~\cite{mmhuman3d}. We train DIMR with batch size 128, using Adam~\cite{kingma2015adam} optimizer with a learning rate of $2.5e^{-4}$ and batch size 128, we set $\lambda_{S M P L}=1$, $\lambda_{3 D}=5$, $\lambda_{2 D}=5$, and $\lambda_{sim}=1$. We take the HRNet-W32 pretrained on COCO poses estimation~\cite{coco} as the backbone. We first train our model without parallel inpainting for 50K iterations, with random scale, rotation, flipping, and channel noise augmentation. Then we finetune the model in parallel training for 15K iterations with synthetic occlusion.

\section{Experiments}

\noindent\textbf{Datasets}: We train DIMR on model using the standard training sets of Human3.6M\cite{h36m}, COCO\cite{coco}, and MuCo\cite{mehta2018single}. We use the ground truth SMPL parameters of Human3.6M and MuCo, and the pseudo ground truth SMPL parameters provided by EFT\cite{eft} for COCO. Our evaluation includes multiple human pose datasets. To assess the general performance of the model, we have evaluated it on 3DPW~\cite{3dpw}. Moreover, we have evaluated the occlusion handling capability of our model on various subsets of 3DPW, namely 3DPW-OCC, which predominantly contains object occlusion, 3DPW-PC, a person occlusion subset of 3DPW, and 3DOH, an object occlusion dataset. For qualitative assessment, we have employed OCHuman\cite{zhang2019pose2seg}, a person occlusion dataset. 

\noindent\textbf{IUV Generation.} We separately train and test our model on two different sets of IUV images. First we use DensePose \cite{Guler2018DensePose} to generate the segmentation and predicted IUV images. To assess our model's full capabilities, we use SMPL parameters to render ground-truth IUV images. Noted that we mask out the invisible portion with a segmentation map, which means the invisible parts won't have any valid dense correspondence. The second model is denoated as DIMR (GT IUV).

\begin{table*}[ht]
    \centering
    \scalebox{1}{\begin{tabular}{c|cccccccc}
\toprule
\small
\multirow{2}{*}{Methods}    &\multicolumn{3}{c}{3DPW-PC} & \multicolumn{3}{c}{3DPW-OC} & \multicolumn{2}{c}{3DOH} \\ 
\cline{2-9} 
& PA-MPJPE$\downarrow$   & MPJPE$\downarrow$  & PVE$\downarrow$    & PA-MPJPE$\downarrow$   & MPJPE$\downarrow$  & PVE$\downarrow$    & PA-MPJPE$\downarrow$     & MPJPE$\downarrow$     \\
\midrule
HMR-EFT~\cite{hmr} & -          & -      & -      & 60.9       & 94.9   & 111.3  & 66.2         & 101.9      \\
SPIN~\cite{spin}   & 82.6       & 129.6  & 157.6  & 60.8       & 95.6   & 121.6  & 68.3         & 104.3     \\
PyMAF~\cite{pymaf} & 81.3       & 126.7  & 154.3  & -          & -      & -      & -            & -         \\
ROMP~\cite{romp}     & 79.7       & 119.7  & 152.8  & 65.9          & -      & -      & -            & -         \\
PARE~\cite{pare}   & -          & -      & -      & 56.6       & 90.5   & 107.9  & 57.1         & 88.6      \\
OCHMR~\cite{ochmr} & 77.1       & 117.5  & 149.6  & -          & -      & -      & -            & -         \\
DIMR (Ours)    & \textbf{70.7}       & \textbf{109.2}  & 154.6  &  \textbf{56.6}          &  \textbf{88.6}      &   108.7     &   58.8           & 91.7           \\
DIMR (GT IUV)   & 54.6      & 88.9  & 104.3  & 50.6       & 75.2   & 94.9   & 50.9         & 79.3      \\ 
\bottomrule
\end{tabular}
}
    \vspace{0.05in}
    \caption{\textbf{Evaluation on occlusion datasets 3DPW-PC\cite{ochmr,3dpw}, 3DPW-OC\cite{pare,3dpw} and 3DOH\cite{ooh}}. For a fair comparison, we train our models without 3DPW datasets when evaluate on both of the 3DPW subset. And we train our models without 3DOH train set when evaluating on 3DOH test set. }
    \label{tab:sota-occlusion}
\end{table*}

\noindent\textbf{Evaluation metrics.} Our evaluation metrics on the 3D datasets include mean per joint position error (MPJPE), Procrustes-aligned MPJPE (PA-MPJPE), and per-vertex error (PVE) in $m m$. MPJPE and PA-MPJPE assess the accuracy of 3D joint rotation, and PVE evaluates the 3D surface error.

\subsection{Comparison to the state-of-the-art}

\textbf{Occlusion Evaluation}: Table\ref{tab:sota-occlusion} compares DIMR's robustness against occlusion with other SOTA occlusion handling methods. We compare our models on occlusion datasets. On 3DPW-PC datasets, our model achieves state-of-the-art performance on 3DPW-OCC dataset. And it has comparable results on other occlusion datasets. On the 3DPW-PC dataset, our model exhibits better performance than OCHMR, a heatmap-based method designed for inter-person occlusion, with a higher PA-MPJPE score. This suggests that our feature extraction method is more accurate compared to the center heatmap-based method, indicating that in terms of accurate pose prediction, our dense map-based method performs better. And comparing to another dense correspondence method PyMAF\cite{pymaf}, our model is clearly better, as a dedicated network for map prediction is more robust under human occlusion. On the 3DOH dataset, our model demonstrates comparable performance with the state-of-the-art method PARE\cite{pare}. The ground truth model achieves a substantial increase in accuracy, as evidenced by a 50\% lower MPJPE on the 3DPW-OCC dataset compared to PARE. On the PA-PMJPE of 3DPW-PC, our model increases performance by 20\%. This dramatic increase in performance on person-occlusion subsets demonstrates that with an accurate dense correspondence map, our network can accurately separate visible human features from occluded people.

\textbf{General Comparison: }We compare our model on 3DPW, which is a general dataset. Our results indicate that DIMR, achieves comparable performance when compared to state-of-the-art methods. Furthermore, we observe that our model outperforms occlusion-specific methods such as OCHMR and ROMP, with a lower error rate. We attribute this success to our model's ability to incorporate both global and local information through our designed architectures, which contribute to the high baseline performance of our model.

\begin{figure*}[ht]
  \centering
\includegraphics[width=0.9\linewidth]{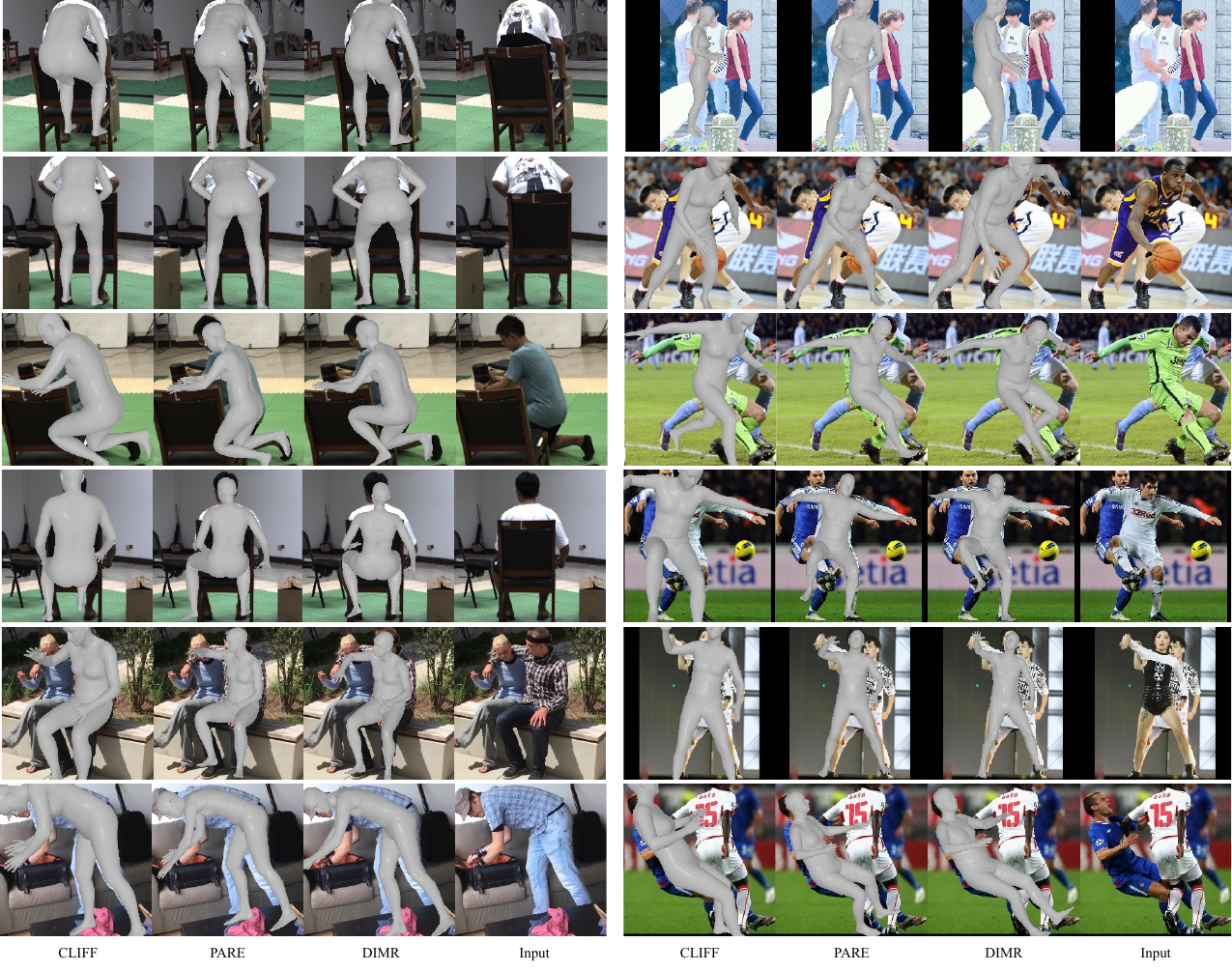}
  \caption{\textbf{Qualitative results on 3DOH (left column 1-4), 3DPW (left column 5-6), and OCHuman (right columns) datasets.} DIMR demonstrates higher accuracy in detecting limbs under various types of occlusion. In instances where human subjects heavily overlap, our model can effectively predict the correct limbs with the assistance of a dense correspondence map.}
  \label{fig:sota}
\end{figure*}

\textbf{Qualitative Comparison:} Figure~\ref{fig:sota} visualizes the qualitative results compared with CLIFF\cite{cliff} and PARE\cite{pare}.We visualize our result on OCHUman\cite{zhang2019pose2seg}, 3DPW\cite{3dpw} and 3DOH\cite{ooh} datasets. DIMR predicts accurate human limbs under object occlusions. Under hard cases like heavy human occlusion, our model is able to predict the correct human with the corresponding limbs, due to the accurate feature separation based on the dense map.
\begin{table}[]
\begin{tabular}{cc|ccc}
\toprule

\multicolumn{2}{c|}{\multirow{2}{*}{Methods}}   & \multicolumn{3}{c}{3DPW}   \\ 
                            \cline{3-5} 
         &                 & PA-MPJPE$\downarrow$ & MPJPE$\downarrow$ & PVE$\downarrow$  \\ 
\midrule

\multirow{6}{*}{\rotatebox{90}{General}}
                        & DecoMR~\cite{decomr}                  & 61.7     & -     & -     \\
                        & SPIN~\cite{spin}                    & 59.2     & 96.9  & 116.4  \\
                        & PyMAF~\cite{pifu}                   & 58.9     & 92.8  & 110.1 \\
                        & HMR-EFT~\cite{eft}                 & 52.2     & 85.1  & 98.7\\
                        & METRO~\cite{metro}                   & 47.9     & 77.1  & 88.2 \\

\hline

\multirow{4}{*}{\rotatebox{90}{Occludsion}}  
                                & OCHMR~\cite{ochmr}                   & 58.3     & 89.7  & 107.1\\
                                & Pose2UV~\cite{pose2uv}                 & 57.1     & -     & -  \\
                                & ROMP~\cite{romp}                    & 53.3     & 85.5  & 103.1 \\
                                & PARE~\cite{pare}                    & 50.9     & 82    & 97.9 \\
                                
\hline
& DIMR (Ours)                    & 52.9     & 87    & 106.6 \\
& DIMR (GT IUV)                 & 47.8    & 73.0 & 99.5 \\
\bottomrule

\end{tabular}
\vspace{0.05in}
\caption{\textbf{Quantitative comparison results} with state-of-the-art methods on 3DPW. We compare the performance of our method with competing for general methods and occlusion-aware methods. } 
\end{table}

\begin{figure}[!t]
  \centering
       

    \centering
    \includegraphics[width=0.9\linewidth]{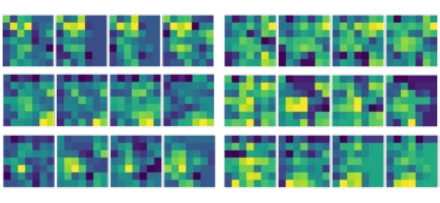}
    \caption{\textbf{Attention map visualization:} we visualize 6 different part-attention layers (left 3 columns and right 3 columns), each representing attention for a human part. We show 4 samples for each part. It shows that our attention focus on relatively the same area (yellow colored) across the instances.}
    \label{fig:attentionmap}
    
\end{figure}
\subsection{Ablation Study}
We conduct our ablation study on 3DPW and 3DPW-OC. In the ablation study, we mainly utilize the ground truth IUV setting for comparison.
\begin{table}[]
\begin{tabular}{c|ccc}
\toprule
\multicolumn{1}{c}{\multirow{2}{*}{Settings}} & \multicolumn{3}{|c}{3DPW}   \\ \cline{2-4} 
\multicolumn{1}{c|}{}                          & MPJPE$\downarrow$  & PA-MPJPE$\downarrow$ & PVE$\downarrow$    \\ 

\midrule
   
   w/o UV feature                              & 103.4 & 59.2    & 129.5  \\ 
   w/o UV structure                            & 80.8  & 51.4    & 106.1 \\ 
   w/o neighboring UV                          & 75.0  & 49.6    & 103.1 \\ 
        DIMR                                        & 73.0 & 47.8    & 99.5  \\ 
   \bottomrule
\end{tabular}
\vspace{0.05in}
\caption{\textbf{Ablation study on different input feature choices.} We compare three settings that remove one of our design elements: feature wrapping, UV structure, and neighboring UV mapping.}
  \label{tab:wrap}
\end{table}

\begin{figure}[h]
  \centering
       

    \centering
    \includegraphics[width=1\linewidth]{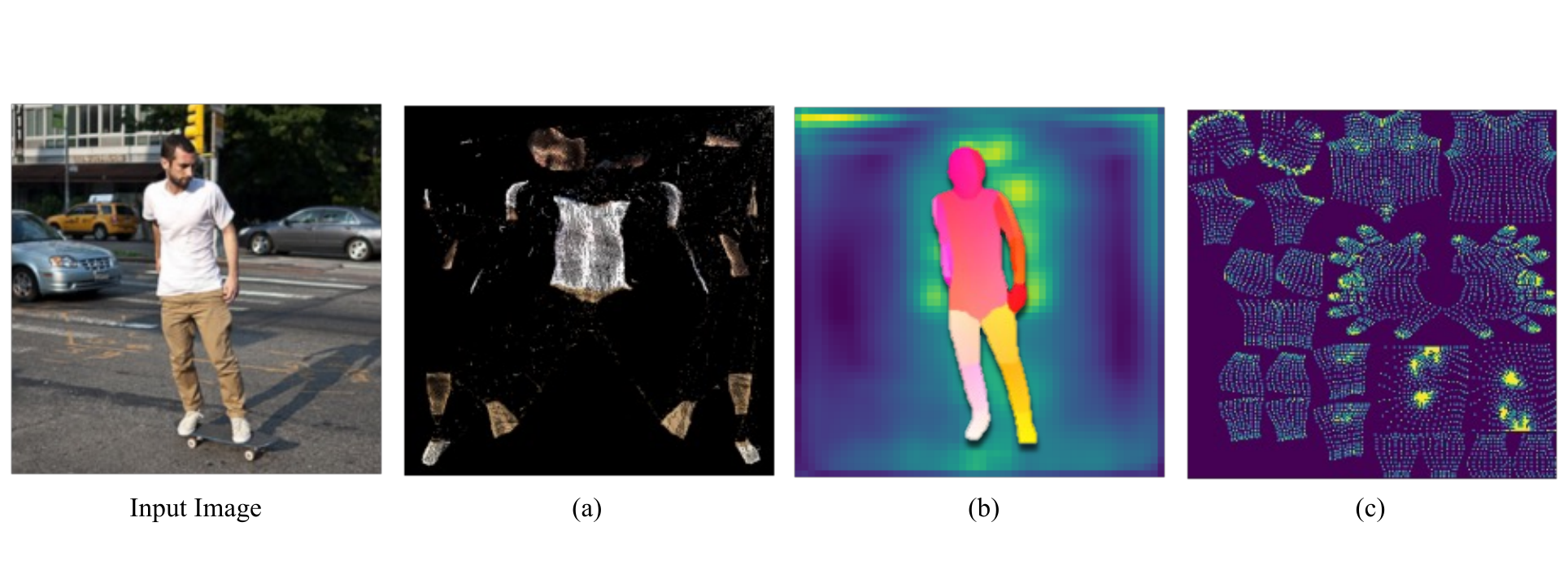}
    \caption{\textbf{The illustration of our modified inputs for the ablation study.} (a) replaces the wrapped feature $F_{UV}$ with the wrapped input image $I_{UV}$; (b) removes inverse-wrapped feature and instead concatenate the $M_{IUV}$ with the image feature map $F_{img}$; (c) replaces the UV map with a randomly organized UV map.}
    \label{fig:iuv_wrapping}
    
\end{figure}

\textbf{Effect of UV feature wrapping:} We first explore the effect of our feature wrapping technique by modifying the input to our feature completion module. The IUV image provides the network with strong prior knowledge, including dense human information. And the wrapping process provides the network with a piece of structured feature information. To assess the individual value of these features, we designe three alternative methods that remove one of these features. For the Table~\ref{tab:wrap}, we substituted wrapped feature $F_{UV}$ with the wrapped input image $I_{UV}$, by wrapping $I$ with  $M_{IUV}$. This setting mainly removes the features of the UV map. For the Table~\ref{tab:wrap}-(b) setting, we remove the inverse-wrapped feature and instead concatenate the $M_{IUV}$ with the image feature map $F_{img}$. We perform feature completion on the original feature map. This setting removes the structured UV map representation but preserves the image feature and dense correspondence. And in the Table~\ref{tab:wrap}-(c) setting, we replace the UV map with a randomly organized UV map, which contains less structured neighboring information. The input feature of these three settings is illustrated in Figure~\ref{fig:iuv_wrapping}. Comparing to the original DIMR, setting (a) cause the most decrease in performance, proving the model require local information for accurate regression. Setting (b) shows limited performance even if it utilizes the ground truth IUV image. At last, setting (c) brings a considerable increase in performance compared to (b), indicating the effectiveness of UV wrapping. But still, the neighboring UV brings more feature fusion.

 \textbf{Effect of wrapped feature completion: }We further perform analysis on the design of our feature completion module. We compare our proposed design with two other settings. In the first setting, we remove the attention layer and only use the down-sampled $F^d_{UV}$ to for the regression, thereby eliminating long-range feature connection. We further replace the convolution down-sample layer with an average-pooling layer, which removes the feature extraction within the body part. Table~\ref{tab:atten} shows our results of different settings. The setting combined convolution layer and attention layer achieves the best results, suggesting that both the local and long-range feature fusion are critical for optimal performance. We also visualize six different part attention layers, each with four instances, shown in Figure~\ref{fig:attentionmap}. Our attention mechanism effectively attended to similar areas across instances for a given body part, which is desirable since the part feature's location remains fixed on the wrapped feature map. Furthermore, it attended to different areas depending on the instance, indicating that it is adaptive. Our attention map also focused on other areas to complete the feature, demonstrating the effectiveness of our attention design.

\begin{table}[]

\small
\setlength\tabcolsep{4pt}
\scalebox{1.1}{\begin{tabular}{c|ccc}
\toprule
        \multirow{2}{*}{Settings} & \multicolumn{3}{c}{3DPW} \\ 
        \cline{2-4}
        & MPJPE$\downarrow$  & PA-MPJPE$\downarrow$  & PVE$\downarrow$  \\ 
\midrule
DIMR     & \textbf{73.0}      & \textbf{47.8}        & \textbf{99.5}    \\
DIMR w/o atten. & 79.6      & 51.1        & 107.6    \\
DIMR w/o atten. conv. & 87.3      & 53.3        & 110.8    \\ 
\bottomrule
\end{tabular}}

\vspace{0.08in}
\caption{\textbf{Ablation study on the input feature.} We compare our original setting with the model without attention and the model without attention and convolution.}
  \label{tab:atten}
\end{table}




\begin{table}[t]
\centering
\scalebox{0.9}{\begin{tabular}{cccc}
\toprule
{Dataset} &                     Settings &                 PA-MPJPE$\downarrow$&     MPJPE$\downarrow$   \\ 

\midrule

\multirow{2}{*}{3DPW-OC}&      synthetic + inpaint &      \textbf{53.4} &        \textbf{82.3} \\

                         &            synthetic     &      53.4 &        84.2\\

\hline                         
\multirow{2}{*}{3DPW}   &      synthetic + inpaint &       50.4 &       \textbf{79.7}  \\

                        &            synthetic     &       \textbf{49.0} &       81.7\\
\bottomrule
\end{tabular}}
\vspace{0.08in}

\caption{\textbf{Ablation study on inpainting training} We compare the models performance with synthetic occlusion and inpainint training on 3DPW, 3DPW-OC.}
\label{tab:inpaint}
\end{table}

\textbf{Effect of wrapped feature inpainting: } In this experiment, we compare our model's ability to handle occlusion in the different training settings. We compare models trained with inpainting training and synthetic occlusion with models with only synthetic occlusion. As shown in Table~\ref{tab:inpaint}, the model utilizing inpainting bring has better accuracy, proving that intermediate supervision is crucial for synthetic occlusion augmentation.


\section{Conclusion}
We present DIMR, a novel human mesh recovery method robust to object and human occlusions. DIMR draws inspiration from the existing methods tackling occlusion. It combines two fundamental ideas: separating the human feature from occlusion and completing features from invisible cues. We leverage the accurate IUV map to isolate the part features of the target individual from those of other humans. The attention-based feature completion module is then utilized to effectively recover the occluded features using cues from other visible parts. In addition, we adopt an inpainting training strategy to enhance performance in comparison to simple synthetic occlusion. Quantitative and qualitative results prove that our model achieves state-of-the-art performance among occlusion-handling methods and has comparable results on general datasets. Future works can focus on the design of occlusion-robust IUV prediction modules for more accurate human mesh recovery under occlusion.


\appendix

\section{More details on Synthetic Occlusion}

\begin{figure}[t]
  \centering
    \centering
    \includegraphics[width=1\linewidth]{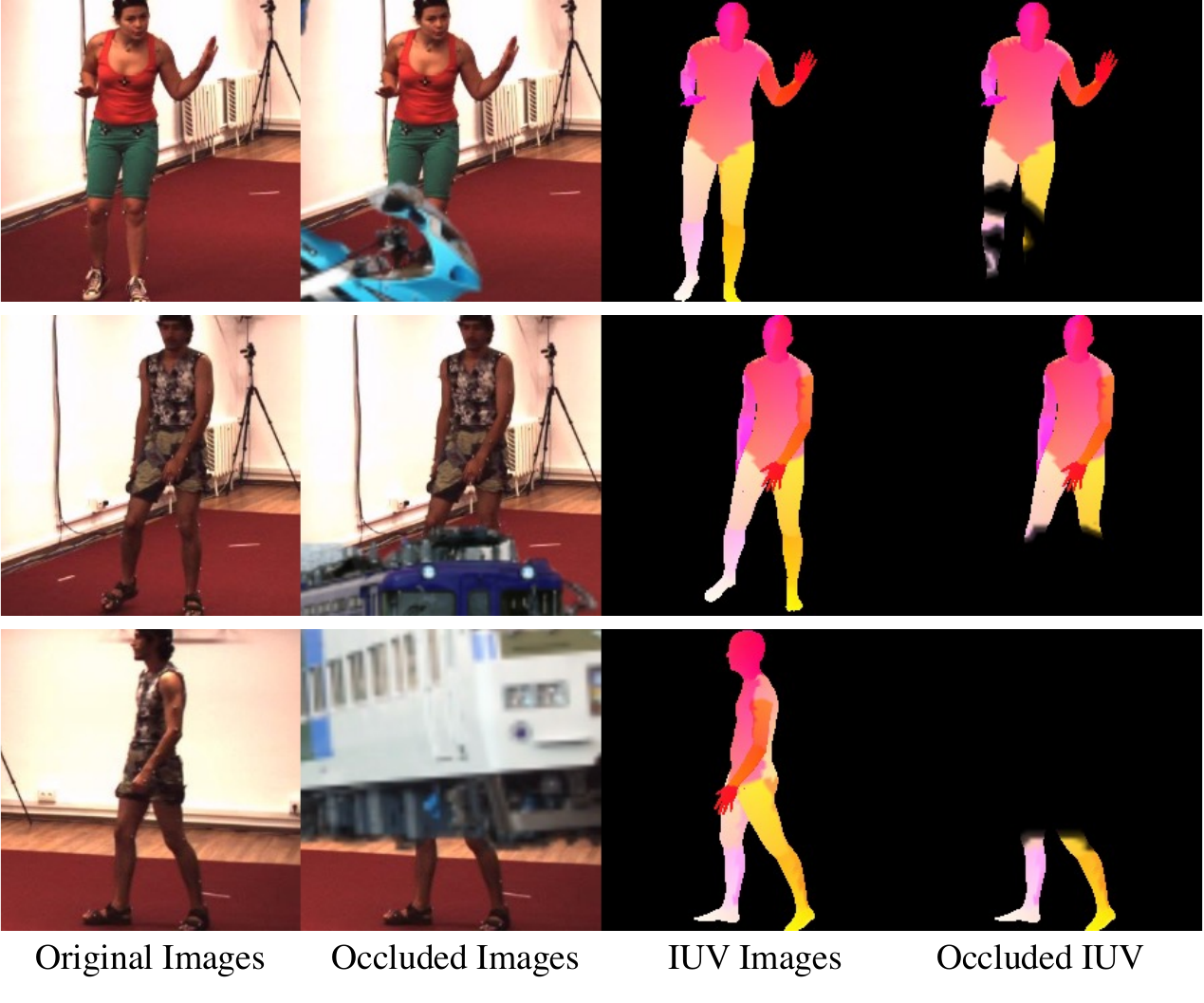}
    \caption{\textbf{Illustration of synthetic occlusion.} We place the objects randomly around the center of the image and randomly resize the objects. The IUV image will also be occluded by the objects.}
    \label{fig:occlu}
    
\end{figure}

We implement the synthetic occlusion following the implementation of PARE~\cite{pare}. We place objects in Pascal VOC~\cite{everingham2009pascal} dataset in a random location around the image center and at a random size. We derive the segmentation mask based on the object and mask out the IUV image with the mask. Figure~\ref{fig:occlu} illustrates the occluded IUV image, the image and the original image. It creates various occlusions to the in-door datasets and increase the diversity of simple datasets like Human3.6M~\cite{h36m}.

\section{Failure Cases}
\begin{figure}[t]
  \centering
    \centering
    \includegraphics[width=1\linewidth]{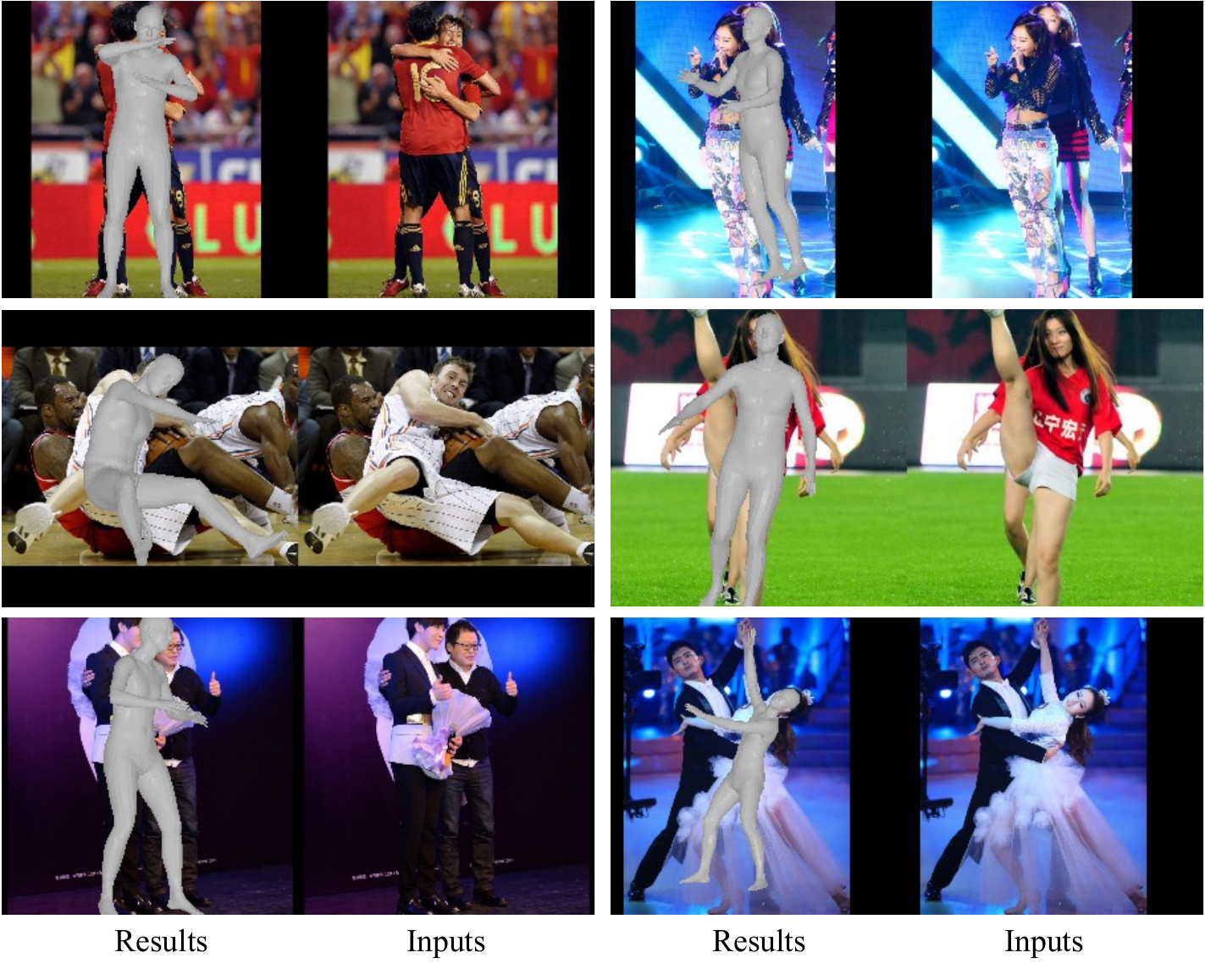}
    \caption{\textbf{Illustration of failure cases.} Top row: failure cases under severe human overlapping. Middle row: failure cases under extreme poses. Bottom row: failure cases under confusing clothes}
    \label{fig:fail}
    
\end{figure}

We present visualizations of inaccurate results on OCHuman~\cite{zhang2019pose2seg} datasets, as demonstrated in Figure~\ref{fig:fail}. The model is currently unable to effectively handle extreme overlapping, such as a hug, where the similarity between limbs and occlusion of the torso presents challenges to accurate prediction. Additionally, limitations in the training data have resulted in restricted performance when dealing with extreme poses, such as laying down or dance poses. Finally, the presence of dark or dress-like clothing can also contribute to inaccurate predictions. These results highlight the need for further research and development to address these challenges.

\section{More Details on IUV Generation}
\begin{figure*}[ht]
  \centering
    \includegraphics[width=1\linewidth]{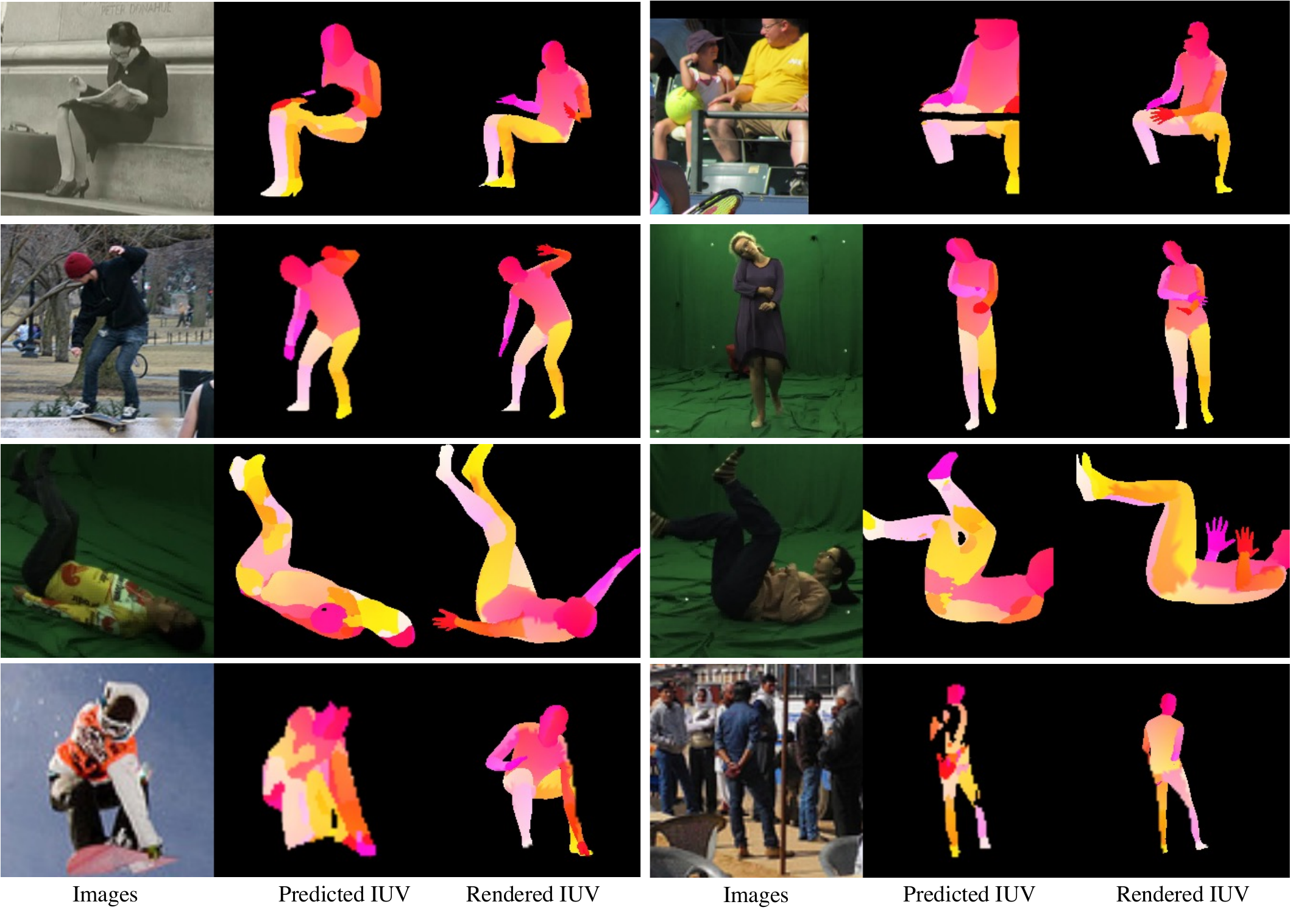}
  \caption{\textbf{IUV images comparison.} Top row shows the predicted IUV images preserve occlusion and body shape information better The second row shows accurate predictions when the body is high resolution. The third row shows inaccurate predictions under extreme poses. And the bottom row shows how the image is}
  \label{fig:iuv_comp}
\end{figure*}
We employ two different methods to generate the IUV images. Firstly, we use the DensePose~\cite{Guler2018DensePose, wu2019detectron2} to predict IUV for all instances in our training and evaluation datasets. For datasets containing single-human instances such as Human3.6M~\cite{h36m}, we directly visualize the predictions of the whole image to generate the IUV image. However, for datasets with multiple human instances, such as MuCo~\cite{singleshotmultiperson2018} and COCO~\cite{coco}, we select the instances based on the intersection over union (IoU) between the predicted and ground truth bounding boxes. We then visualize the predicted IUV image. 

We render the IUV images from the ground-truth SMPL model in the second method. For datasets that provide segmentation maps like COCO~\cite{coco}, OCHuman~\cite{zhang2019pose2seg}, and 3DOH~\cite{ooh}, we overlay the instance segmentation mask. And for other datasets~\cite{3dpw}, we use Mask-RCNN~\cite{matterport_maskrcnn_2017} to generate instance masks.

In Figure~\ref{fig:iuv_comp}, we present the ground truth IUV image, the rendered IUV image, and the predicted IUV image for comparison. The predicted IUV image preserves occlusion information better than the rendered IUV image. The quality of the predicted IUV image is found to be comparable to the rendered image when the human instance is clear. Furthermore, the predicted IUV image is more accurate in terms of body shape. However, the quality of prediction is found to be less desirable under extreme poses or when the image is blurred. Since the IUV prediction model is trained only on COCO~\cite{coco} dataset, it might be less accurate on out-of-domain data.

{\small
\bibliographystyle{ieee_fullname}
\bibliography{egbib}
}


\end{document}